\documentclass{article}
\usepackage{arxiv}
\usepackage[utf8]{inputenc} 
\usepackage[T1]{fontenc}    
\usepackage{xurl}
\usepackage[hidelinks,breaklinks]{hyperref}
\usepackage{url}            
\usepackage{booktabs}       
\usepackage{amsmath,amssymb,amsfonts}
\usepackage{nicefrac}       
\usepackage{microtype}      
\usepackage{graphicx}
\usepackage{float}
\usepackage{threeparttable}
\usepackage{tabularx}
\usepackage{adjustbox}
\usepackage{subcaption}
\usepackage{natbib}
\usepackage{doi}

\title{Imagery Dataset for Remaining Useful Life Estimation of Synthetic Fibre Ropes}

\author{ \href{https://orcid.org/0000-0000-0000-0000}{\includegraphics[scale=0.06]{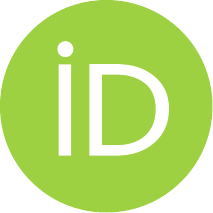}\hspace{1mm}Anju Rani}\thanks{Use footnote for providing further
		information about author (webpage, alternative
		address)---\emph{not} for acknowledging funding agencies.} \\
	Department of Energy\\
	Aalborg University\\
	Esbjerg, Denmark 6700 \\
	\texttt{aran@energy.aau.dk} \\
	\And
	\href{https://orcid.org/0000-0000-0000-0000}{\includegraphics[scale=0.06]{orcid.pdf}\hspace{1mm}Daniel Ortiz-Arroyo} \\
	Department of Energy\\
	Aalborg University\\
	Esbjerg, Denmark 6700 \\
	\texttt{doa@energy.aau.dk} \\
	\And
	\href{https://orcid.org/0000-0000-0000-0000}{\includegraphics[scale=0.06]{orcid.pdf}\hspace{1mm}Petar Durdevic} \\
	Department of Energy\\
	Aalborg University\\
	Esbjerg, Denmark 6700 \\
	\texttt{pdl@energy.aau.dk} \\
}

\date{}

\renewcommand{\shorttitle}{Imagery Dataset for Remaining Useful Life Estimation of Synthetic Fibre Ropes}

\hypersetup{
pdftitle={A template for the arxiv style},
pdfsubject={q-bio.NC, q-bio.QM},
pdfauthor={David S.~Hippocampus, Elias D.~Striatum},
pdfkeywords={First keyword, Second keyword, More},
}

\begin{document}
\maketitle

\begin{abstract}
Remaining useful life (RUL) estimation of synthetic fibre ropes (SFRs) is critical for safe operation in offshore-crane, wind turbine installation, and heavy-load handling applications, where rope failure can result in catastrophic safety incidents and costly downtime. Despite growing research interest in data-driven condition monitoring, there is no publicly available image dataset that captures the complete degradation lifecycle of SFRs under controlled cyclic fatigue loading. To address this gap, we present a novel image dataset comprising approximately 34,700 high-resolution images of eleven Dyneema SK75/78 high-modulus polyethylene (HMPE) rope samples subjected to cyclic fatigue on a sheave-bend test stand at seven distinct axial load levels ranging from 60 kN to 280 kN. Ropes were loaded until mechanical failure, with fatigue lifetimes ranging from 695 cycles to 8,340 cycles. After every fixed number of sheave cycles (an inspection burst), ten images were captured at different cross-sectional positions along the rope, providing spatially representative sampling of surface degradation throughout the rope's entire service life. The images obtained from each load are annotated with the corresponding elapsed cycle count, enabling a direct computation of RUL for any rope in the sequence. This dataset aims to support a broad range of machine learning (ML) tasks including RUL regression, damage progression modelling, anomaly detection, and load-conditioned prognostics. The dataset is intended to serve as a benchmark resource for the development and comparison of vision-based condition monitoring (CM) and prognostics algorithms for SFRs.
\end{abstract}

\keywords{Synthetic fibre ropes \and Condition monitoring \and Defect detection \and Remaining useful life \and Computer vision}

\section*{Specifications Table} 
\begin{tabular}{p{4cm}p{11cm}}  
\hline
Subject                        & Computer Vision and Pattern Recognition \\
Specific subject area          & Vision-based Remaining Useful Life (RUL) estimation of synthetic fibre ropes (SFRs) under cyclic fatigue loading  \\
Type of data                   & Image (PNG)  \\
How data were acquired         & Images were acquired using a Basler acA2000 high-speed camera fitted with a Basler C11-5020-12M-P 12-megapixel lens, operating at 165 FPS at a resolution of 1920 × 1200 pixels. Aputure AL-MC RGBWW LED lights are used to provide illumination on the respective rope. An NVIDIA Jetson Nano P3450 managed image capture. Ten images of different cross-sectional positions along the rope were captured at the end of each inspection burst (a fixed number of sheave cycles), providing spatially representative coverage of the rope surface. \\
Data format                    & Raw (PNG)   \\
Description of data collection & Eleven Dyneema SK75/78 HMPE rope specimens (8 mm nominal diameter, 12-strand braided) were subjected to cyclic fatigue on a sheave-bend test stand. Seven axial load levels ranging from 60 kN to 280 kN were applied. Ten images of different cross-sectional positions along the rope were captured at the end of every inspection burst, as detailed in Table~\ref{tab:rope_data}. Testing continued until each rope failed. Total fatigue lives ranged from 695 cycles (at 250 kN) to 8,340 cycles (at 60 kN). The complete dataset contains approximately 34,700 images across all 11 ropes.\\
Data source location           & \begin{tabular}[c]{@{}l@{}}
                                Institution: Department of Energy, Aalborg University\\ 
                                City/Region: Esbjerg\\ 
                                Country: Denmark \\
                                \end{tabular}  \\
Data accessibility             & \begin{tabular}[c]{@{}l@{}}
                                Repository name: Kaggle \\ 
                                DOI: \url{10.34740/kaggle/dsv/16105762}\\ 
                                Direct URL to data: \href{https://www.kaggle.com/datasets/anu6942/imagery-dataset-for-rul-estimation-of-fibre-ropes}{Imagery Dataset for RUL Estimation of Fibre Ropes} \\ 
                                \end{tabular} \\ 
\hline
\end{tabular}

\section*{Value of the Data}
\begin{itemize}
\item SFRs made from Dyneema (HMPE) are widely deployed in offshore crane systems, wind turbine installation vessels, and heavy lifting operations. Fatigue-induced failure in such applications can be catastrophic. The present work provides the first publicly available lifecycle imagery dataset for HMPE ropes, enabling the development of automated, vision-based remaining useful life (RUL) prediction systems that can provide early warning of impending failure, supporting timely rope replacement and reducing unplanned maintenance costs.
\item This imagery dataset captures the complete visual degradation trajectory of SFRs under controlled cyclic sheave-bend fatigue loading, from healthy condition through progressive degradation to failure. Covering eleven rope specimens across seven distinct load levels (60-280 kN) with fatigue lives ranging from 695 to 8,340 cycles and approximately 34,700 total images, it provides a statistically meaningful and load-diverse benchmark not available in any existing dataset.
\item The multi-load-level design of the dataset enables researchers to study the influence of applied tension on fatigue degradation rates and visual appearance, supporting the development of load-conditioned RUL prediction models. The ten spatially distributed images per inspection burst further capture the non-uniform spatial nature of fatigue damage along the rope length, providing rich training signal than single-view acquisition strategies.
\item Each rope used in the test is annotated with its elapsed cycle count and associated rope identity, enabling straightforward derivation of RUL labels for supervised regression, classification into life-stage bins (early, mid, late), or unsupervised anomaly detection. The dataset is compatible with a wide range of deep learning (DL) architectures including CNNs, Vision Transformers (ViT), and multimodal vision-language models (VLMs), supporting diverse experimental paradigms.
\item This dataset complements the author's previously published defect classification dataset \cite{rani2023imagerydatasetconditionmonitoring}, which focused on identifying damage type and severity. Together, these two resources form a comprehensive imagery benchmark covering both damage identification and lifecycle-level prognostics for Dyneema SFRs, addressing the full condition monitoring (CM) pipeline from fault detection to remaining life prediction.
\end{itemize}

\begin{figure}[ht]
     \centering
     \includegraphics[width=\textwidth]{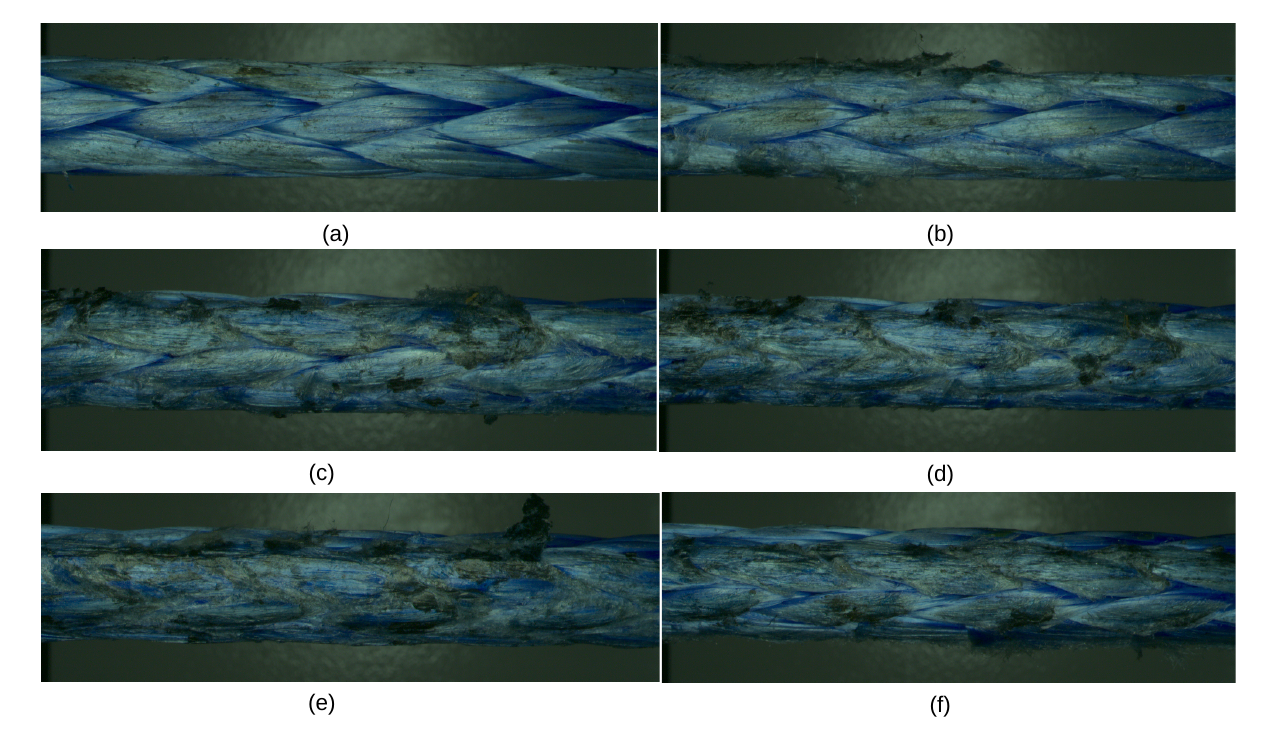}
     \caption{Lifecycle images for rope specimens from healthy condition (a) to failure (f), illustrating the accelerating rate of visual degradation.}
\label{fig:RUL}
\end{figure}

\section{Objective}
Synthetic fibre ropes, particularly those manufactured from high-modulus polyethylene (HMPE) fibres such as Dyneema, have become increasingly prevalent in offshore and heavy-load applications as lightweight, high-strength alternatives to conventional steel wire ropes (SWRs) \cite{mckenna2004handbook}. HMPE offers high strength, immunity to corrosion, low density, and torsion-neutral braided construction, making them especially well suited for offshore crane hoisting systems, wind turbine installation vessels, and deep-sea mooring \cite{RANI2024104109}. 
Despite these advantages, HMPE ropes are susceptible to progressive fatigue degradation during cyclic bending-over-sheave (BOS) operations. Repeated contact and bending against sheave surfaces generate internal fibre-on-fibre abrasion, inter-strand compressive loading, and surface wear, progressively reducing the rope's load-bearing capacity and ultimately leading to mechanical failure. Because fatigue damage accumulates gradually and is not spatially uniform along the rope length, accurate estimation of the RUL is a challenging but operationally critical task \cite{ridge2001effect}.
Current industry practice relies primarily on periodic manual visual inspection supplemented by retire-on-time policies, both of which are conservative, subjective, and poorly suited to the variable loading conditions encountered in real operations. Data-driven, vision-based condition monitoring offers a compelling pathway to continuous, objective, and quantitative RUL estimation; however, progress in this direction has been impeded by the absence of suitable publicly available datasets.
Our previous work introduced a defect-classification imagery dataset for Dyneema ropes, enabling the identification of damage types and severity levels \cite{rani2023imagerydatasetconditionmonitoring}. While that dataset supports fault detection tasks, it does not capture temporal lifecycle information and therefore cannot directly support RUL estimation. To address this gap, we present a publicly available imagery dataset comprising approximately 34,700 images of eleven HMPE rope specimens captured throughout their complete service life - from healthy rope to mechanical failure - under controlled sheave-bend cyclic fatigue at seven load levels between 60 kN and 280 kN. This dataset is intended to serve as the primary benchmark dataset for the development, training, and evaluation of vision-based RUL estimation and prognostic health management (PHM) algorithms for SFRs.

\begin{table}[h!]
\centering
\begin{tabular}{lccccc}
\hline
\textbf{Rope} & \textbf{Load} & \textbf{Images} & \textbf{Cycles/burst} & \textbf{Total cycles} \\
\hline \hline
Rope1  & 100 kN & 3,746 & 20 & 7,492 \\
Rope2  & 60 kN  & 5,175 & 15 & 7,762 \\
Rope3  & 200 kN & 2,100 & 10 & 2,100 \\
Rope4  & 150 kN & 3,579 & 10 & 3,579 \\
Rope5  & 150 kN & 3,547 & 10 & 3,547 \\
Rope6  & 100 kN & 1,710 & 10 & 1,710 \\
Rope7  & 60 kN  & 8,340 & 10 & 8,340 \\
Rope8  & 250 kN & 695   & 10 & 695 \\
Rope9  & 150 kN & 3,638 & 10 & 3,638 \\
Rope10 & 280 kN & 722   & 10 & 722 \\
Rope11 & 220 kN & 1,415 & 10 & 1,415 \\
\hline
\textbf{Total} & 60-280 kN & ~34,700 & 10-20 & --\\
\hline
\end{tabular}
\caption{Synthetic fibre rope fatigue experiments across 7 axial load levels (60 - 280 kN).}
\label{tab:rope_data}
\end{table}

\section{Data Description}
The present dataset consists of approximately 34,700 high-resolution PNG images acquired from eleven Dyneema SK75/78 HMPE rope specimens subjected to cyclic fatigue on a sheave-bend test stand. Images have a resolution of 1920 × 1200 pixels and were captured using a Basler acA2000 camera. The rope images were captured at different cross-sectional positions along the rope at the end of every inspection burst. The burst interval (cycles-per-burst) was set in the range of 10–20 sheave cycles, depending on the applied load level, with shorter intervals used at higher loads to provide denser temporal sampling of the accelerated degradation.
Data collection was done in periodic inspection bursts. After the completion of the sheave cycles (the cycles-per-burst), the test stand paused rotation and ten images were captured of different cross-sectional positions along the rope length. This spatial sampling strategy was adopted because fatigue damage in SFRs is not spatially uniform: local contact geometry, braid construction irregularities, and loading eccentricities cause damage to develop at different rates at different positions along the rope. Ten images per burst provide a spatially representative snapshot of rope condition at each elapsed cycle count.
All ten images within an inspection burst share the same elapsed cycle label and the same RUL value (computed as the total failure cycle count minus the elapsed cycle count). Testing continued until each rope reached complete mechanical failure, defined as a sudden loss of load-bearing capacity. The dataset provides complete lifecycle coverage for all eleven specimens.
The dataset is organised into eleven rope folders, one per specimen. Each folder contains all images acquired from the first inspection burst to the burst immediately preceding failure, along with a metadata file recording the experimental parameters (load level, rope speed, LCS, cycles-per-burst, and total image count). Table~\ref{tab:rope_data} summarises the eleven rope specimens, their applied load levels, total fatigue life, and approximate image counts. The seven distinct load levels spanning 60 kN to 280 kN provide a wide dynamic range of fatigue behaviour, enabling load-conditioned modelling and cross-load generalisation studies.

\begin{figure}[ht]
\centering
\includegraphics[width=\textwidth]{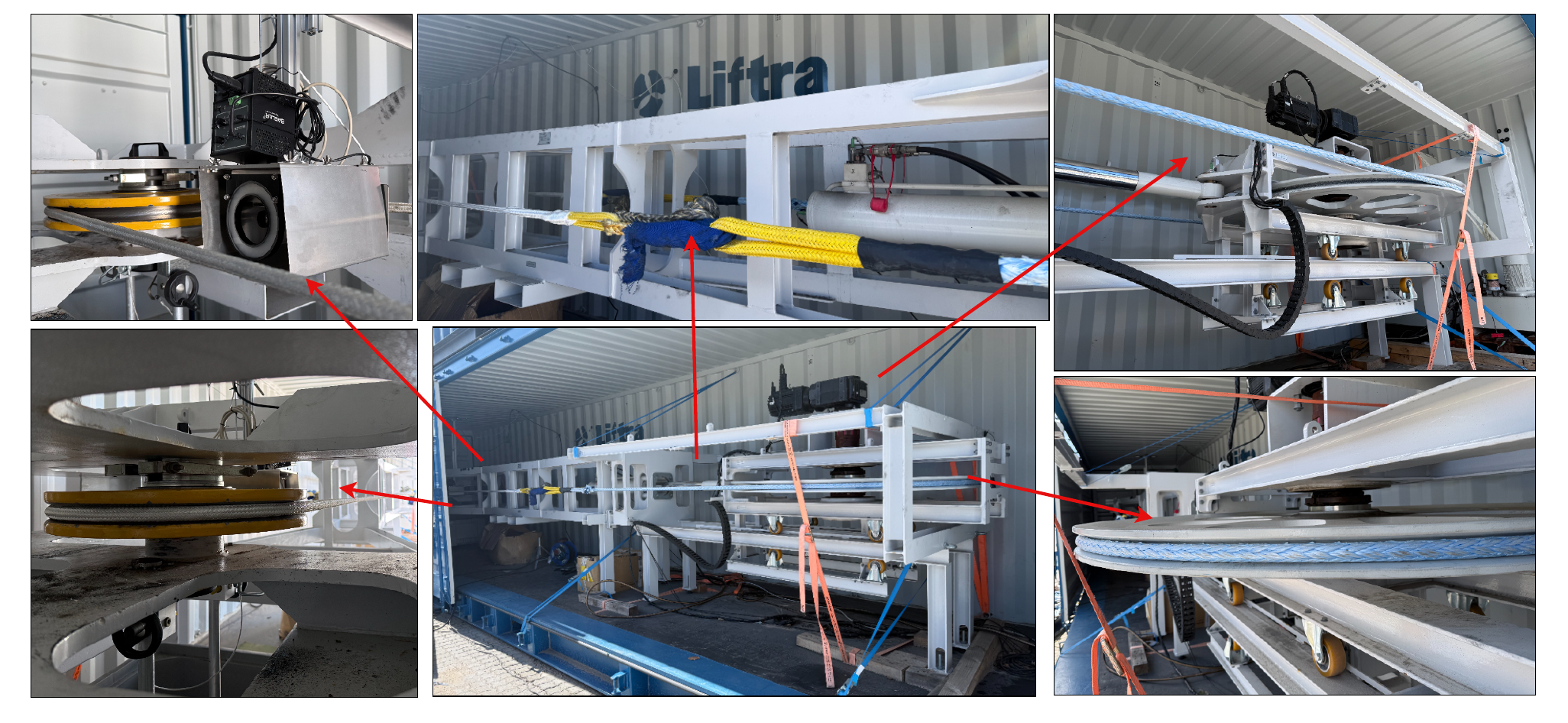}
\caption{Sheave-bend fatigue test stand (\cite{liftra}, \cite{dynamica}) used for cyclic loading of Dyneema SK75/78 HMPE rope specimens. The central image shows the full test frame with the rope specimen installed under axial tension. Surrounding close-up views highlight: (top-left and bottom-left) the camera unit and sheave assembly showing the rope bending contact zone; (top-centre) the rope specimen under load with visible colour-coded strand markers; (top-right and bottom-right) the multi-sheave pulley arrangement and rope routing path; (bottom-centre) the complete test stand assembly housed within the test container facility.}
\label{fig:setup}
\end{figure}

\section{Experimental Setup}
Rope specimens were subjected to cyclic fatigue on a sheave-bend test stand. An axial tension load was applied to the rope while the sheave rotated back and forth, bending the rope over the pulley in a repeating cycle. This BOS loading regime replicates the primary failure mechanism experienced by hoisting and crane ropes in operational service. After completion of the selected sheave cycles, rotation stopped, and ten images were taken of different cross-sections along the rope length. Taking ten images at different positions provides a spatially representative sample of rope condition, since fatigue damage is not spatially uniform. These ten images constitute one inspection burst; all images within a burst share the same elapsed cycle count and therefore receive identical RUL labels. Testing continued until each rope failed completely. Illumination was provided by the Aputure AL-MC RGBWW LED lights positioned at approximately 45° to the rope surface to provide uniform, shadow-free lighting across the rope cross-section during image capture. Image acquisition was managed by an NVIDIA Jetson Nano P3450 embedded computing platform, which interfaced with the camera system and triggered burst capture at the specified cycle intervals.

\subsection{Imaging System}
Images were acquired using a Basler acA2000 GigE Vision camera connected via Ethernet, fitted with a Basler C11-5020-12M-P Premium 12-megapixel lens at a resolution of 1920 × 1200 pixels. The camera operated in continuous acquisition mode with software triggering at an exposure time of 5000 µs, with auto-exposure disabled to ensure photometric consistency across all images and all rope specimens. Raw frames were captured in BayerRG8 pixel format and converted to BGR8 prior to PNG storage. Strobe synchronisation between the camera exposure and the Aputure AL-MC RGBWW LED lights was achieved by configuring the camera output as a strobe signal, ensuring the lights fired precisely and exclusively during each exposure window, eliminating ambient light variation between frames.
Image capture was coordinated by an NVIDIA Jetson Nano P3450 running a Python acquisition pipeline (pypylon, OpenCV) that monitored the test stand programmable logic controller (PLC) over Ethernet using the pylogix library. At the end of every inspection burst, the PLC raised a TakePicture tag; the acquisition system detected the rising edge, executed a software trigger, converted and saved the image in PNG format with a filename encoding a sequential burst index and a millisecond-precision timestamp, and confirmed successful capture to the PLC via a PictureCompleted handshake tag. The PLC then cleared the trigger before the next burst commenced. This synchronised, handshake-based architecture guaranteed that no inspection burst was missed and that every saved image could be unambiguously associated with its elapsed sheave cycle count, ensuring the integrity of the RUL labelling scheme across the entire dataset.

\subsection{Rope Description}
All eleven rope specimens were manufactured from Dyneema SK75/78, a multi-filament HMPE fibre produced via a gel-spinning process that yields an exceptionally high degree of molecular orientation and crystallinity, resulting in a tensile strength approximately 15 times greater than structural steel on a weight-for-weight basis \cite{mckenna2004handbook}, \cite{dynamica}. The rope specimens share the following construction specification:

\begin{itemize}
\item  Fibres: Dyneema SK 75/78
\item  Nominal Diameter: 8 mm
\item  Construction: 12 strands / 12 braided rope
\item  Torsional neutral: The rope is designed in a way to resists twisting or torsional forces.
\item  Pitch/stitch length: Approximately 11mm
\item  Braiding period: Approximately 66mm
\end{itemize}

\subsection{Loading Conditions and Data Collection}
Seven distinct axial load levels were applied across the eleven rope specimens, ranging from 60 kN to 280 kN. This range was selected to span the operational loading envelope typical of offshore crane and lifting applications and to produce a spectrum of fatigue life suitable for benchmarking RUL algorithms across a wide range of degradation rates. Observed fatigue life ranged from 695 sheave cycles at 250 kN (Rope8) to 8,340 sheave cycles at 60 kN (Rope7), reflecting the expected inverse relationship between applied load and fatigue life.
Failure was defined as a sudden, complete loss of load-bearing capacity in the rope specimen, resulting in separation of the rope. This objective criterion ensured consistent definition of the failure endpoint across all eleven specimens, which is essential for the accurate computation of RUL labels from elapsed cycle counts.

Each rope specimen was subjected to cyclic tensile loading under the following fixed experimental conditions:
\begin{itemize}
\item 	Cylinder force: Range between 60 kN - 280 kN
\item 	Rope speed: 60 m/min
\item 	LCS length: 1000 mm
\item 	Cycles to Camera Routine: Range between 10 - 20 (e.g. ten image sets captured every 20 loading cycles)
\item 	Camera: Basler acA2000 with Basler C11-5020-12M-P Premium 12-megapixel lens
\item 	Frame rate: The camera operated in sequential software-triggered single-frame acquisition mode. Image capture was initiated by the test stand PLC at the end of each inspection burst; no fixed frame rate was applied. 
\item 	Image resolution: 1920 × 1200 pixels
\item 	Image format: PNG
\end{itemize}

\subsection{RUL Labelling}
Each rope folder in the dataset is labelled with the elapsed cycle count after the images are captured. The RUL for any image is computed as:

\begin{equation}
\mathrm{RUL} = N_{\text{failure}} - N_{\text{elapsed}}
\end{equation}

where $N_{\text{failure}}$  is the total sheave cycle count at mechanical failure for the rope specimen, and $N_{\text{elapsed}}$ is the cycle count at which the image was captured. This labelling scheme supports both continuous RUL regression (predicting the exact remaining cycle count) and discrete life-stage classification (e.g., early life: RUL > 60\% of total life; mid life: 30\%–60\%; late life: < 30\%).

\subsection{Compliance with Standards}
The experimental design and loading conditions were developed with reference to ISO 9554:2019 (Fibre ropes - General specifications) \cite{ISO9554} and BOS fatigue testing guidelines for synthetic fibre ropes. These standards informed the selection of loading parameters, rope construction specifications, and failure criteria applied throughout the study.

\section*{Ethics Statement}
This research does not involve experiments, observations, or data collection related to human or animal subjects. All experiments were conducted on synthetic fibre rope specimens in a controlled laboratory environment.

\section*{Declaration of competing interest}
The authors declare that they have no known competing financial interests or personal relationships that could have appeared to influence the work reported in this paper.

\section*{Data Availability}
\href{https://www.kaggle.com/datasets/anu6942/imagery-dataset-for-rul-estimation-of-fibre-ropes} {Imagery Dataset for RUL Estimation of Fibre Ropes}. 

\section*{Acknowledgement}
This research was supported by Aalborg University, Liftra ApS (Liftra), and Dynamica Ropes ApS (Dynamica) in Denmark under the EUDP program through project grant number 64021-2048.

\bibliographystyle{unsrtnat}
\bibliography{references}  
\end{document}